\documentclass{article}

\usepackage{iclr2027_conference,times}

\usepackage{amsmath,amsfonts,bm}

\def\eqref#1{equation~\ref{#1}}
\def\1{\bm{1}}

\DeclareMathAlphabet{\mathsfit}{\encodingdefault}{\sfdefault}{m}{sl}
\SetMathAlphabet{\mathsfit}{bold}{\encodingdefault}{\sfdefault}{bx}{n}

\usepackage{url}
\usepackage{amssymb}
\usepackage{amsthm}
\usepackage{mathtools}
\usepackage{empheq}
\usepackage{subcaption}
\usepackage{wrapfig}
\usepackage{graphicx}
\usepackage{xspace}
\usepackage{algorithm}
\usepackage[noend]{algorithmic}
\usepackage{booktabs}
\usepackage{bbm}
\usepackage{enumitem}
\usepackage{bm}
\usepackage{blindtext}
\usepackage{multicol}
\usepackage{multirow}
\usepackage{arydshln}

\usepackage{hyperref}
\hypersetup{
    colorlinks=true,
    linkcolor=blue,
    citecolor=brown,
    urlcolor=blue
}

\def\method{Anchor-WM\xspace}

\title{
Pretrained Features as Teachers, Not States:
Compact Latent World Models for Visual Planning
}

\author{Yanbo Wang\enskip\enskip
Nan Lin\enskip\enskip
Wendelin Böhmer\enskip\enskip
Javier Alonso-Mora  \\
Delft University of Technology (TU Delft) \\
}

\iclrfinalcopy

\begin{document}

\maketitle

% Remove ICLR publication header
\thispagestyle{fancy}
\fancyhead{}

% Add preprint note on first page
\begingroup
\renewcommand{\thefootnote}{}
\footnotetext{Preprint.}
\endgroup
\setcounter{footnote}{0}

% Your abstract and paper continue here

\maketitle

\begin{abstract}

A key design choice in visual world models is what information the latent state is encouraged to preserve. Models that learn a latent space with only a latent-prediction objective adapt to environment dynamics, but the predictive objective alone admits collapsed solutions and provides no explicit constraint on the visual information retained by the latent. Models built directly on pretrained visual features inherit a strong visual prior, but tie prediction and planning to a frozen state space. We propose Anchor-WM, which instead uses pretrained visual features as a teacher for learning a compact latent state rather than as the world-model state itself. Anchor-WM learns this compact latent with action-conditioned dynamics, while distilling the dense DINO representations through a training-only distillation head. This encourages the compact latent to retain visual information captured by the pretrained representation while discouraging representational collapse. At inference, the DINO encoder and distillation head are discarded, allowing prediction and planning to operate entirely in the learned latent space. Anchor-WM achieves the best average success rate across the four evaluated environments while enabling substantially faster planning than world models operating directly on DINO patch features. These results show that pretrained visual features can effectively guide world-model learning without defining the state space used for prediction and planning.
%\href{https://universal-simulator.github.io}{https://universal-simulator.github.io}.

%1. Pretrained visual features can be more effective as a training signal for a world model than as its state space.allowing the world model to plan in a latent two orders of magnitude smaller at comparable success.

%2. Unlike a generic distributional regularizer, this constraint is observation-dependent: it encourages the latent to preserve visual information captured by the pretrained representation rather than imposing a prescribed marginal geometry on the latent space.

%3 \textbf{Grounding is training-only.} The DINO teacher ($22$M, frozen) and the reconstruction decoder $h_\phi$ ($14.6$M) are used solely to form the training objective and are discarded at inference. The deployed world model is a $16.4$M-parameter network (ViT-tiny encoder $5.5$M + AR predictor $10.8$M + action encoder $0.16$M), so distillation adds \emph{training} cost but leaves inference and planning cost unchanged.

\end{abstract}

\section{Introduction}

World models aim to capture how an environment evolves over time conditioned on an agent's actions so as to support prediction, planning, and control~\citep{ha2018world}. In visual domains, they often rely on learned latent states to represent observations compactly, and a central challenge is determining what information these states should retain. A common approach is to learn such representations with observation reconstruction as part of the training objective~\citep{hafner2020dream,hafner2023dreamerv3}, but a reconstruction objective requires modeling fine-grained visual details, such as textures and lighting variations, even when these details are irrelevant to downstream control.

One alternative is to model dynamics in a pretrained visual representation rather than in pixel space. DINO-WM learns a world model that operates directly on pretrained DINO patch features, providing a spatially structured visual representation for offline model-based control and planning~\citep{zhou2025dinowm}. While this avoids predicting raw observations and provides a strong visual prior enabling zero-shot generalization, directly adopting the frozen pretrained representation as the world-model state imposes two limitations. First, the state cannot adapt to the dynamics and control requirements of the target environment. Second, DINO represents an image as a spatial grid of patch features, yielding a dense state that must be propagated through the dynamics model for every candidate action sequence during planning, which can make planning computationally expensive.

A different approach is to learn the world-model state end-to-end from pixels, while predicting future states in latent space rather than reconstructing pixels, following the Joint Embedding Predictive Architecture (JEPA)~\citep{lecun2022path}. Recent JEPA-based world models, such as LeWM, take this route~\citep{maes2026leworldmodel,terver2026ebjepa}, allowing the learned state to freely adapt to the target environment rather than inheriting a fixed pretrained state space. In reward-free settings, however, the predictive objective alone does not determine which visual information the representation should preserve and, in the extreme, admits collapsed solutions. These methods therefore rely on architectural or regularization strategies to prevent collapse. Avoiding collapse, however, does not by itself determine which visual information the learned state should retain.

These approaches differ in how they define the world-model state. Using a pretrained visual representation directly provides a semantically rich state, but fixes the representation on which dynamics and planning must operate. Learning the state end-to-end allows it to adapt to the task and dynamics, but provides no explicit constraint on what visual information it should preserve. This raises a natural question: \textit{Can pretrained visual representations instead guide the learning of a compact predictive state, without being used directly as the state for prediction and planning?}

We propose \method, which uses pretrained visual features only as training-time supervision to learn a compact latent space for prediction and planning. The compact latent is trained jointly to predict future latent states conditioned on actions and to reconstruct dense DINO features through a lightweight distillation head. This encourages the learned latent to retain visual information captured by DINO while remaining free to adapt to the environment dynamics. The distillation objective also helps prevent representational collapse by providing an additional constraint on the learned latent space. At inference, the DINO encoder and distillation head are discarded, and prediction and planning operate entirely in the learned compact latent space instead of dense DINO feature space. %Our formulation therefore decouples the pretrained representation used for supervision from the compact latent state used for prediction and planning.

%Unlike feature-space alignment methods that make the deployed representation directly resemble the pretrained feature space, Anchor-WM only requires pretrained features to be decodable from the learned state, leaving the state free to acquire a compact geometry suited to dynamics and planning.

We evaluate \method across diverse visual control environments against multiple world-model baselines including DINO-WM and LeWM. \method achieves the highest average success rate across the considered environments, while remaining competitive with the best-performing method on each individual task, suggesting that it benefits from the visual prior provided by pretrained features while retaining the flexibility to adapt its latent state to the target dynamics. Meanwhile, \method plans approximately $100\times$ faster than DINO-WM because planning is performed entirely in a compact learned latent space rather than directly in the high-dimensional DINO feature space, with planning speed comparable to LeWM. These results show that pretrained visual features can guide the learning of a compact latent state that supports strong planning performance without incurring the inference-time cost of operating directly in the pretrained feature space. %At the same time, the performance variation across environments highlights that information useful for visual representation learning does not necessarily coincide with all of the fine-grained information required for dynamics and control.

Overall, we contribute the following: (1) We introduce \method, a world model that uses pretrained visual features as training-time guidance while performing prediction and planning entirely in a compact latent space; (2) We develop a feature-distillation objective that distills pretrained visual information into the latent space while helping prevent representational collapse; and (3) We demonstrate across diverse visual control environments that this design yields strong planning performance with substantially lower planning cost than operating directly in pretrained feature space.

\section{Related Work}
\label{sec:related_work}

\paragraph{Latent World Models.}
Many world-modeling approaches learn dynamics in latent spaces. PlaNet~\citep{hafner2019planet}, Dreamer~\citep{hafner2023dreamerv3}, and IRIS~\citep{micheli2023transformers} use observation reconstruction as part of representation learning, whereas TD-MPC~\citep{hansen2024tdmpc2} learns a decoder-free latent model shaped by reward and value supervision. JEPA-based world models instead learn representations by predicting future latent without reconstructing pixels or predicting rewards
\citep{lecun2022path,assran2023ijepa,bardes2024vjepa}.  A central challenge when learning such representations end-to-end is representational collapse, which prior work addresses through target encoders or explicit representation regularization
\citep{bardes2021vicreg,bardes2024vjepa,sobal2022joint,sobal2025stress,assran2023ijepa,balestriero2025lejepa,maes2026leworldmodel,terver2026ebjepa}. Recent work further constrains the information or structure captured by the learned state through action-based objectives \citep{ivashkov2026sensorimotor,zhang2026deltajepa} or temporal relationships derived from offline trajectories \citep{bai2026tdjepa}. Our proposed approach constrains the latent representation learning through distilling DINO features, encouraging the learned state to retain visual information captured by DINO while directly discouraging representational collapse.

\paragraph{Pretrained Visual Representations for Control and World Modeling.}
%Large-scale visual pretraining provides representations that transfer across perception and embodied-control tasks. Self-supervised vision models such as DINO and DINOv2 learn general-purpose visual features from unlabeled images \citep{caron2021dino,oquab2023dinov2}, while R3M, MVP, VIP, and related approaches investigate visual pretraining specifically for robotic and embodied control \citep{nair2023r3m,xiao2022mvp,ma2023vip,radosavovic2023realworld}. These methods demonstrate that pretrained visual features can provide useful perceptual representations for downstream control without learning perception entirely from task-specific interaction.
Large-scale visual pretraining provides representations that transfer across
perception and embodied-control tasks. Self-supervised models such as DINO and DINOv2 learn general-purpose visual features \citep{caron2021dino,oquab2023dinov2}, while R3M, MVP, VIP, and related work study visual pretraining specifically for robotic control
\citep{nair2023r3m,xiao2022mvp,ma2023vip,radosavovic2023realworld}. Pretrained representations have also been incorporated into world models. DINO-WM operates directly on frozen DINOv2 patch features, predicting future spatial representations from offline trajectories and planning toward visual goals in the DINO feature space \citep{zhou2025dinowm}. V-JEPA 2 leverages large-scale self-supervised video pretraining and post-trains an action-conditioned latent world model for robot planning \citep{assran2025vjepa2}. Temporal Straightening learns planning-oriented latent representations by projecting pretrained DINO features through a lightweight projector and applying a curvature regularizer to shape the latent geometry~\citep{wang2026temporal}. Across these pretrained-feature approaches, the pretrained representation remains part of the inference procedure---as the state space, as the pretrained backbone, or as the input to a learned projector. Our approach instead uses the pretrained representation only as a training target, so neither the pretrained encoder nor its representation is required for prediction or planning.

\paragraph{Pretrained Representations as Teachers.}
Our approach is also related to knowledge distillation, which transfers information from a teacher to a separately learned student, including through intermediate representations and their relational structure \citep{hinton2015distilling,romero2015fitnets,zagoruyko2017attention,ahn2019variational,tung2019similarity}. More recently, pretrained visual representations have provided auxiliary supervision for models optimized for other objectives. REPA aligns projected diffusion-transformer representations with frozen pretrained visual features \citep{yu2025repa}, a principle extended to multimodal world models by $M^2$-REPA \citep{m2repa} and to action grounding by AGRA \citep{agra2026making}. Related work has also distilled pretrained visual features into compact descriptors for matching and retrieval, where the compressed descriptor is itself the deployed representation \citep{lee2021distill}. Our approach uses pretrained feature distillation to learn a compact world-model state. Rather than matching the teacher's representation space, a training-only distillation head encourages the dense DINO representation to remain recoverable from a single compact latent. The teacher thus guides representation  learning without defining the state space used for prediction and planning.

\section{Method}
\label{sec:method}

\begin{figure}[!t]
    \centering
    \includegraphics[width=\textwidth]{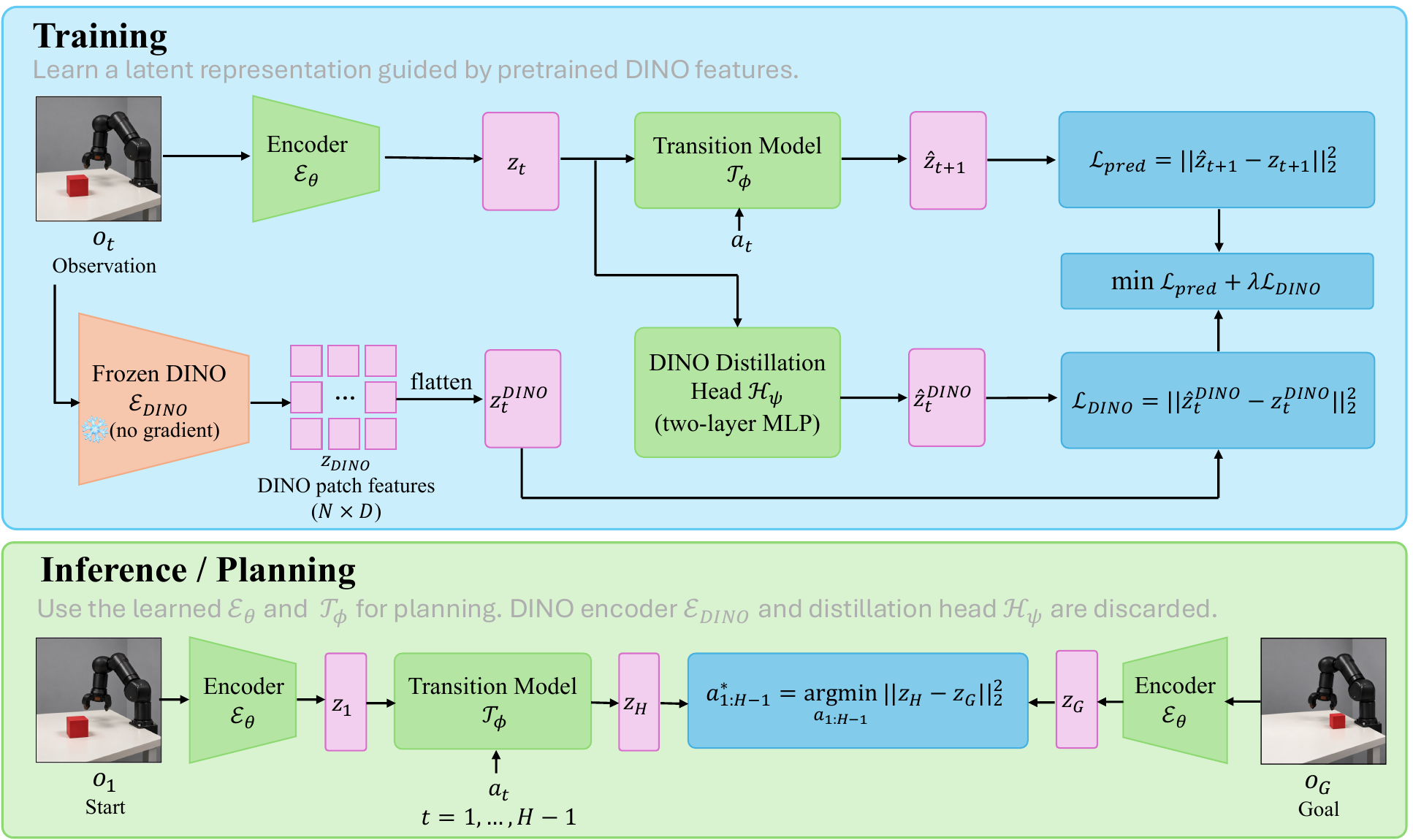}
    \vspace{-1mm}
    \caption{\textbf{Training and inference of Anchor-WM.} During training, the observation encoder $\mathcal{E}_\theta$, transition model $\mathcal{T}_\phi$, and distillation head $\mathcal{H}_\psi$ are optimized jointly, while the pretrained DINO encoder remains frozen. The compact latent is trained with both a dynamics prediction objective and a distillation objective that reconstructs DINO features. At inference, the DINO encoder and distillation head are discarded, and planning is performed entirely in the learned latent space by optimizing an action sequence whose predicted terminal latent approaches the encoded goal latent.}
    \label{fig:overview}
\end{figure}

%We introduce \method, a visual world model that learns a compact latent state for prediction and planning. We first outline the architecture of the predictive world model. We then show how pretrained DINO features can be used solely as a training-time guidance signal. Finally, we describe how the learned dynamics are used for planning entirely within the compact latent space. An overview of \method is shown in Fig.~\ref{fig:overview}.

%We introduce Anchor-WM, a world model that learns a latent state for prediction and planning by using pretrained DINO features as training-time guidance. 

We consider an offline visual control setting with a dataset of $M$ trajectories $\mathcal{D}=\{\tau^{(i)}\}_{i=1}^{M}$, where $\tau^{(i)}=(o_0,a_0,o_1,a_1,\ldots,o_T)$, $o_t$ denotes an image observation, and $a_t$ the action taken at time $t$. Our goal is to learn a compact
latent dynamics model from these offline trajectories that supports both
prediction and visual planning. We next introduce \method.

\subsection{Latent World Model}
\label{sec:predictive_latent}

To learn a latent world model, \method uses an encoder $\mathcal{E}_\theta(\cdot)$ to map an image observation $o_t \in \mathbb{R}^{H \times W \times C}$ to a compact latent
state $z_t \in \mathbb{R}^{d}$:
\begin{equation}
z_t = \mathcal{E}_\theta(o_t).
\label{eq:encoder}
\end{equation}
Dynamics are modeled directly in this compact latent space. Given a context
of $K$ latent states $z_{t-K+1:t}$ and actions $a_{t-K+1:t}$, \method uses a transition
model $\mathcal{T}_\phi$ to predict the next latent state as
\begin{equation}
\hat{z}_{t+1}
=
\mathcal{T}_\phi
\left(
z_{t-K+1:t},
a_{t-K+1:t}
\right).
\label{eq:dynamics}
\end{equation}
%At planning time only observations up to $t$ are available, so the context window slides over previously predicted latents as the rollout proceeds.
%We instantiate the encoder and transition model using the lightweight architecture of LeWM \citep{maes2026leworldmodel}. The encoder $\mathcal{E}_\theta$ is a vision transformer, ViT-tiny \citep{dosovitskiy2020image}, and we use its final-layer \texttt{[CLS]} token directly as $z_t$. LeWM applies an additional MLP with batch normalization to the \texttt{[CLS]} embedding as part of its anti-collapse regularization. We omit this projection and instead shape the latent through DINO feature distillation (\S\ref{sec:dino_guidance}). The transition model $\mathcal{T}_\phi$ is a Transformer, with actions incorporated at each layer through AdaLN \citep{peebles2023dit}. 
We train the encoder and transition model by minimizing the mean squared error between the predicted and target next latent states:
\begin{equation}
\mathcal{L}_{\mathrm{pred}}
=
\mathbb{E}_{(o_{t-K+1:t},
a_{t-K+1:t},o_{t+1})\sim \mathcal{D}} \left\|
\hat{z}_{t+1} - z_{t+1}
\right\|_2^2.
\label{eq:predictive_loss}
\end{equation}

%Predicting in latent space avoids requiring the model to reproduce pixel-level details that need not be useful for modeling future dynamics. However, the predictive objective alone provides limited guidance on what visual information should be retained in the learned representation and may admit collapsed solutions without additional constraints \citep{maes2026leworldmodel}. We therefore guide latent learning using features from a pretrained visual encoder. Since these features vary across observations, reconstructing them discourages representational collapse while providing the latent with a rich visual learning signal.

Training the world model with only this predictive objective admits a degenerate solution in which the encoder maps every observation to a constant latent and the transition model predicts the same constant. We therefore introduce DINO-guided latent learning to constrain this degeneracy while providing a pretrained visual learning signal.

\subsection{DINO-Guided Latent Learning}
\label{sec:dino_guidance}

We guide the latent representation learning using features from a frozen pretrained DINOv2 encoder $\mathcal{E}_{\mathrm{DINO}}(\cdot)$. For each observation $o_t$, we extract the post-LayerNorm DINO patch features $\mathcal{E}_{\mathrm{DINO}}(o_t)\in \mathbb{R}^{N\times D}$ and flatten them in their fixed spatial order to form the distillation target $z^{\text{DINO}}_t \in \mathbb{R}^{ND}$, where $N$ is the number of patches, and $D$ is the embedding dimension. We retain the full patch-level representation rather than pooling it into a global target
%, encouraging the single-token latent, i.e., the student, 
to retain spatially localized visual information. 
%Importantly, the learned state need not share the spatial tokenization of the pretrained representation: a single compact latent is supervised to recover the full ordered patch representation. 
We validate this design choice against global and patch-wise distillation variants in Sec.~\ref{sec:ablations}.  %We also validate we regress the post-LayerNorm DINO features (not raw), reducing norm-domination; an L2-vs-cosine ablation is in Sec.~\ref{sec:ablations}.

Rather than using $z^{\mathrm{DINO}}_t$ as the world-model state like DINO-WM, we use it only as a target for representation learning. We introduce a lightweight two-layer MLP distillation head with GELU activation, $\mathcal{H}_\psi: \mathbb{R}^{d} \rightarrow \mathbb{R}^{ND}$, which maps the compact latent $z_t$ to the flattened DINO target space:
\begin{equation}
    \hat{z}^{\text{DINO}}_t = \mathcal{H}_\psi(z_t).
    \label{distillation}
\end{equation}
This design confines the high-dimensional patch representation to the
training objective: the world model itself continues to predict and plan
using a single compact latent state. %We ablate global and patch-wise alternatives in Sec.~\ref{sec:ablations}.

We apply the distillation in Eq~\ref{distillation} to every observation in the sampled context window. Thus, each latent state is trained to regress onto the post-LayerNorm DINO patch features of its corresponding observation by minimizing
\begin{equation}
\mathcal L_\text{DINO} = \mathbb E_{o_t \sim \mathcal{D}}\|\mathcal H_\psi(z_t) - z^{\text{DINO}}_t \|_2^2.
\label{eq:dino_loss}
\end{equation}

%\begin{equation}
%\mathcal{L}_{\mathrm{DINO}}
%=
%\sum_{i=t-K+1}^{t}
%\left\|
%\hat{z}^{\text{DINO}}_i - z^{\text{DINO}}_i
%\right\|_2^2.
%\label{eq:dino_loss}
%\end{equation}
%Thus, each latent state is trained to reconstruct the post-LayerNorm DINO patch features of its corresponding observation.

The distillation objective encourages DINO features \(z^{\text{DINO}}_t\) to remain recoverable from the compact latent \(z_t\) without directly constraining \(z_t\) to match \(z^{\text{DINO}}_t\). A learnable distillation head maps \(z_t\) into the DINO feature space, allowing \(z_t\) to have a different dimensionality and internal structure as long as \(z^{\text{DINO}}_t\) can be predicted from it. This objective also discourages representational collapse: if the encoder mapped every observation to the same latent, the distillation head would receive the same input for all observations and could only produce a constant prediction. Therefore, minimizing the distillation objective requires the encoder to produce observation-dependent latents.

\subsection{Training \method}

To train \method, we jointly optimize the predictive and distillation objectives as
\begin{equation}
\mathcal{L}
=
\mathcal{L}_{\mathrm{pred}}
+
\lambda\mathcal{L}_{\mathrm{DINO}},
\label{eq:total_loss}
\end{equation}
where $\lambda$ controls the strength of the distillation objective. Both objective terms shape the learned encoder $\mathcal{E}_\theta$: prediction encourages the latent to capture information useful for modeling action-conditioned dynamics, while distillation encourages it to retain visual information represented by the pretrained DINO features. The transition model $\mathcal{T}_\phi$ and distillation head $\mathcal{H}_\psi$ are optimized by their respective objectives, while the DINO encoder $\mathcal{E}_{\text{DINO}}$ remains frozen throughout training.

%Crucially, the pretrained DINO features serve only as training targets and do not define the state space of the world model. This decouples the representation used to guide learning from the compact latent used for prediction and planning. At inference, the DINO encoder and distillation head are discarded, so both dynamics prediction and planning operate entirely in the learned latent space.

\subsection{Planning with \method}
\label{sec:planning}

At inference, the encoder $\mathcal{E}_\theta$ maps the observed context into the latent space, and the transition model $\mathcal{T}_\phi$ autoregressively rolls out future latent states under candidate action sequences. Since observations beyond time $t$ are unavailable, the context window is updated with predicted latent states as the rollout proceeds. We select actions by approximately solving
\begin{equation}
a_{t:t+H-1}^{*}
=
\arg\min_{a_{t:t+H-1}}
\mathcal{C}\!\left(\hat{z}_{t+1:t+H}\right),
\label{eq:planning}
\end{equation}
where $\hat{z}_{t+1:t+H}$ is obtained by rolling out $\mathcal{T}_\phi$ and
$\mathcal{C}$ denotes the planning objective.

For the goal-reaching tasks considered in this work, we minimize the distance between the predicted terminal latent $\hat{z}_{t+H}$ and the latent representation of the goal observation $o^{\mathrm{goal}}$:
\begin{equation}
\mathcal{C}\!\left(\hat{z}_{t+1:t+H}\right)
=
\left\|
\hat{z}_{t+H}
-
\mathcal{E}_\theta\!\left(o^{\mathrm{goal}}\right)
\right\|_2^2 .
\label{eq:planning_cost}
\end{equation}
%The planning objective is therefore evaluated directly in the latent space.
%The objective is thus evaluated directly in the learned latent space rather than in the pretrained DINO feature space. Although DINO features guide the learned representation during training, planning does not require explicit comparison in the DINO feature space. Instead, the latent is jointly shaped by the predictive and distillation objectives and is used directly as the state space for goal-conditioned planning.

We approximately solve Eq.~\ref{eq:planning} using the Cross-Entropy Method (CEM) \citep{rubinstein2004cross}. CEM iteratively samples candidate action sequences, evaluates their predicted trajectories, and refines the sampling distribution toward lower-cost candidates. We execute the selected action sequence for the full planning horizon and then replan from the newly
observed state. Planning horizons and CEM hyperparameters are given in Section~\ref{sec:app_model} of the Appendix.

%Planning with \method therefore operates entirely on the compact learned state, whose dimensionality is independent of the pretrained DINO representation. This is particularly advantageous for sampling-based planning, where many candidate action sequences must be repeatedly rolled out through the transition model. A patch-based state requires each candidate rollout to propagate the full token grid, whereas \method\ rolls out a single compact latent. By confining the pretrained representation to training, \method\ retains itsrepresentation-learning signal without carrying its token-wise computational cost into planning. The DINO encoder and distillation head are discarded after training and add no inference cost. We quantify the resulting difference in planning time in Section~\ref{sec:efficiency}.

Because planning rolls out a single compact latent rather than a grid of DINO patch features, its computational cost is independent of the dimensionality of the teacher representation. This is particularly advantageous for sampling-based planning, which repeatedly evaluates many candidate action sequences. We quantify the resulting difference in planning time in Sec~\ref{sec:planning_eval}.

\section{Experiments}

%In this section, we evaluate the planning performance and physical understanding of \method. In Section~\ref{sec:planning_eval}, we evaluate visual planning under both standard and long-horizon settings, and further examine zero-shot generalization to novel environment configurations not seen during training. In Section~\ref{sec:physical_understanding}, we analyze whether \method preserves the physical and temporal information required for visual planning. Finally, in Section~\ref{sec:ablations}, we conduct ablation studies to assess the contributions of the key design choices in our model.
\begin{figure}[!t]
    \centering
    \includegraphics[width=\textwidth]{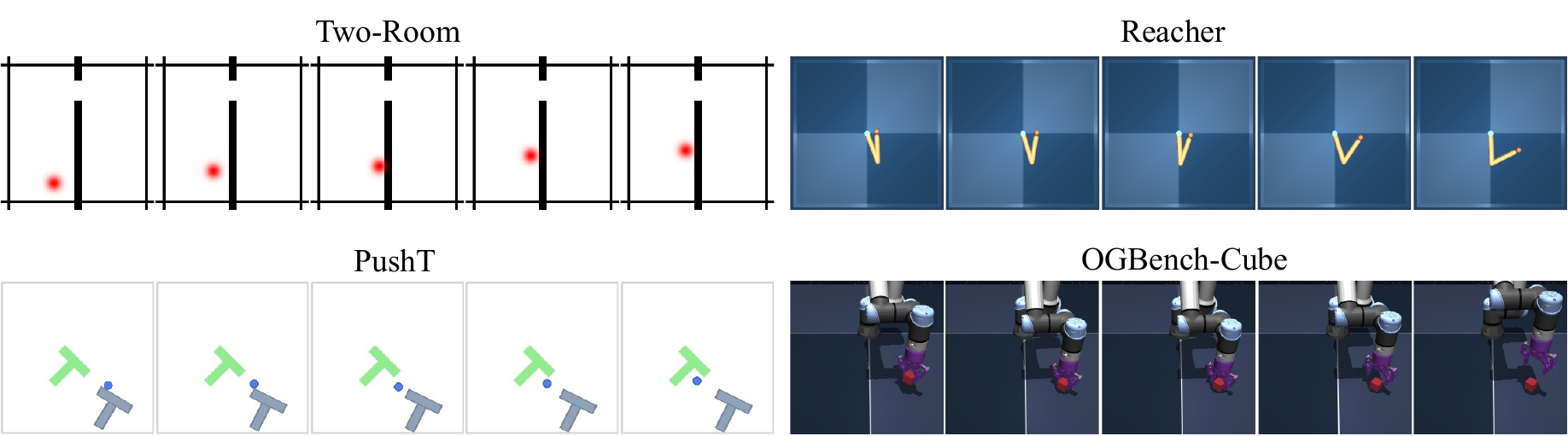}
    \caption{\textbf{Evaluation Environments.} Top left: Two-Room for 2D navigation. Top right: Reacher for planar two-link arm control. Bottom left: Push-T for contact-based 2D object manipulation. Bottom right: Cube for 3D robotic cube manipulation. Further details are given in Section~\ref{sec:app_data}.}
    \label{fig:environments}
    \vspace{-5pt}
\end{figure}
\subsection{Experimental Setup}

\paragraph{Environments.} We evaluate \method on TwoRoom \citep{sobal2025stress}, Reacher \citep{tassa2018deepmind}, Push-T \citep{chi2025diffusion}, and OGBench-Cube \citep{park2025ogbench} as illustrated in Fig.~\ref{fig:environments}, with further details provided in the Appendix.

\paragraph{Baselines.}
We compare \method with several baselines. \textit{Prediction-Only} is a latent world model trained solely with a prediction objective, without any additional regularization, serving as an ablation baseline. \textit{PLDM} \citep{sobal2025stress} and \textit{LeWM} \citep{maes2026leworldmodel} are JEPA-based world models that further employs statistical regularization to mitigate representational collapse. \textit{DINO-WM} \citep{zhou2025dinowm} instead uses frozen DINO patch features, thereby avoiding representation collapse. 
%Temporal Straightening \citep{wang2026temporal}, which we abbreviate as DINO-TS, learns a task-specific representation on top of a frozen DINO encoder and uses stop-gradient in the prediction objective to mitigate representational collapse. 
%The original implementation of \textit{DINO-WM} additionally uses proprioceptive information alongside visual observations to facilitate world-model learning. Since
We focus on learning world models solely from visual observations, without using additional modalities such as proprioceptive inputs. We therefore evaluate all methods under the same visual-only setting to ensure a fair comparison.

 %We additionally include a \textit{Predictive-only} baseline, trained solely with the latent prediction objective and without additional representation regularization, and a \textit{Random} baseline that samples actions uniformly at random.

\paragraph{Architecture.} We instantiate the encoder $\mathcal{E}_\theta$ as a ViT-Tiny (patch size 14, hidden width 192, 12 layers, and 3 attention heads) and the transition model $\mathcal{T}_\phi$ as a Transformer (6 layers, 16 attention heads, hidden width 2048, head dimension 64, and dropout 0.1), with action conditioning via AdaLN following \citep{maes2026leworldmodel}. The Anchor-WM latent state, $z_t\in\mathbb{R}^{192}$, is obtained by projecting the CLS token produced by the ViT-Tiny encoder. To distill DINO features, we use a lightweight two-layer MLP distillation head $\mathcal{H}_\psi$ that predicts the flattened DINOv2 patch features, $\mathbb{R}^{196\times384}$, from $z_t$. Additional architectural details are provided in the Appendix.

\subsection{Planning Evaluation}
\label{sec:planning_eval}

\begin{table*}[!t]
\centering
\caption{
\textbf{Planning Performance Comparison}.
Success Rate ($\uparrow$) across four environments, with the final column reporting the average across tasks. Bold indicates the highest mean success rate in each environment. 
%$^\dagger$ denotes results not statistically distinguishable from the best-performing method.
}
\vspace{-3pt}
\label{tab:success_rate}
\resizebox{\textwidth}{!}{
\begin{tabular}{lcccccc}
\toprule
Method
& Latent Size
& TwoRoom
& Reacher
& PushT
& Cube

& Average \\
\midrule

%Random
%& 1 token
%& 43 
%& 3 / 171
%&8 / 151
%& 14 / 131
%& 76.6 / 121 \\

Predictive
& 1 $\times$ 192
& 35.2$\pm$ 2.2
& 10.6$\pm$ 5.3
& 3.2 $\pm$ 1.1
& 46.5 $\pm$ 2.9
& 23.9 \\

%\midrule

PLDM
& 1 $\times$ 192
& 93.0 $\pm$ 2.0
& 69.5 $\pm$ 5.8
& 71.8 $\pm$ 6.1
& 51.2 $\pm$ 1.6
& 71.4 \\

LeWM
& 1 $\times$ 192
& 86.1 $\pm$ 1.8
& 78.4 $\pm$ 5.7
& \textbf{90.4 $\pm$ 4.3}
& 64.8 $\pm$ 2.1
& 79.9 \\

%\midrule

%TS-global
%& 1 token
%& 10.4 / 1.66
%& 49.2 / 1.29
%& 32.8 / 1.58
%& 68.8 / 1.25
%& 40.3 / 1.45 \\

DINO-WM
& 196 $\times$ 384
& \textbf{99.6 $\pm$ 0.9}
& 67.2 $\pm$ 1.8
& 65.2 $\pm$ 6.6
& \textbf{74.4 $\pm$ 2.6}
& 76.6 \\

DINO-WM (CLS)
& 1$\times$ 384
& 59.3 $\pm$ 1.7
& 45.2 $\pm$ 3.4
& 37.8 $\pm$ 5.6
& 50.2 $\pm$ 3.8
& 48.1 \\

%DINO-TS
%& 196 \times 8
%& 99.6 / 63
%& 59.2 / 105
%& 52.4 / 104
%& 75.6 / 77
%& 71.8 / 87 \\

%\midrule

%\textbf{Ours (distill, proj)}
%& \textbf{1 token}
%& 74.0 / 1.41
%& 71.2 / 1.25
%& \textbf{82.4} / 1.45
%& 99.6 / 1.02
%& \textbf{81.8 / 1.28} \\

%Ours (distill, no-proj)
%& 1 token
%& 84.8 / 1.18
%& $44^{\ddagger}$ / 2.34
%& $74.8^{\rightarrow}$ / 1.44
%& 98.8 / 0.93
%& $(75.6^{\ddagger} / 1.47)$ \\

Anchor-WM (ours)
& 1 $\times$ 192
& \textbf{99.6 $\pm$ 0.9}
& \textbf{82.4 $\pm$ 4.3}
& 84.8 $\pm$ 5.4
& 72.0 $\pm$ 4.2

& \textbf{84.7} \\

\bottomrule
\end{tabular}
\vspace{-5pt}
}\vspace{-10pt}
\end{table*}

\begin{table}[!b]
\centering
\caption{\textbf{Long-horizon planning performance.} Planning success rates when the goal observation is selected 50 timesteps after the start observation and the evaluation budget is 100 steps.}
\vspace{-3pt}
\label{tab:longhorizon}
\begin{tabular}{lccccc}
\toprule
Method & TwoRoom & Reacher & PushT & Cube & Average\\
\midrule
LeWM & 79.2$\pm$3.6 & 95.2$\pm$1.8 & 41.6$\pm$7.1 & 44.8$\pm$8.3 & 65.2\\
DINO-WM & 96.0$\pm$2.5 & 90.0$\pm$5.7 & 32.0$\pm$2.8 & \textbf{53.2$\pm$5.0} & 67.8\\
Anchor-WM (ours) & \textbf{96.4$\pm$1.7} & \textbf{96.4$\pm$3.3} & \textbf{46.4$\pm$3.6} & 45.2$\pm$2.7 & \textbf{71.1}\\
\bottomrule
\end{tabular}
\vspace{-15pt}
\end{table}

\paragraph{Planning Results.}
Table~\ref{tab:success_rate} reports the planning performance of Anchor-WM and the baselines, where the goal is selected 25 time steps after the start state along an expert trajectory, with a planning budget of 50 steps. Anchor-WM achieves the highest success rate on Reacher and ties DINO-WM (patch) for the best performance on TwoRoom, while remaining competitive on PushT and Cube. Although LeWM performs best on PushT and DINO-WM (patch) performs best on Cube, Anchor-WM attains the highest average success rate across all four environments. These results suggest that using DINO features as training-time supervision enables Anchor-WM to learn compact, task-specific representations that support strong visual planning performance across diverse environments.

We also evaluate Anchor-WM and the baselines under longer-horizon planning, where the goal observation is selected 50 timesteps after the start observation and the evaluation budget is increased to 100 steps. Table~\ref{tab:longhorizon} reports the corresponding success rates. Anchor-WM achieves the highest success rate on TwoRoom, Reacher, and PushT, as well as the highest average success rate across all environments, while DINO-WM performs best on Cube. These results suggest that guiding a compact, task-specific latent representation with pretrained DINO supervision remains effective for longer-horizon planning.

\begin{table}[t]
\centering
\caption{
\textbf{Zero-shot Planning Performance.} \method and the baselines are trained on the original environments shown in Fig.~\ref{fig:environments}. At test time, we introduce unseen visual perturbations to evaluate zero-shot generalization, such as changing the end-effector or target-object color.
}
\label{tab:planning_zero_shot}
\begin{tabular}{lccccc}
\toprule
Method & TwoRoom & Reacher & PushT & Cube & Average \\
\midrule
LeWM
& 62.8$\pm$8.2
& 56.4$\pm$8.4
& 10.8$\pm$3.0
& 64.0$\pm$4.0
& 48.5 \\

DINO-WM
& 98.0$\pm$2.4
& \textbf{60.8$\pm$7.0}
& \textbf{62.8$\pm$4.8}
& \textbf{76.4$\pm$3.6}
& \textbf{74.5} \\

Anchor-WM (ours)
& \textbf{99.2$\pm$1.1}
& 47.2$\pm$10.8
& 33.2$\pm$4.8
& 68.0$\pm$6.5
& 61.9 \\
\bottomrule
\end{tabular}
\vspace{-5pt}
\end{table}
We further evaluate the planning performance of \method in the zero-shot setting, where at inference time we introduce visual perturbations that are not observed during training, such as changes in the end-effector or object color. Examples of these perturbations are illustrated in Fig.~\ref{fig:environments_ood}. As shown in Table~\ref{tab:planning_zero_shot}, \method substantially improves over LeWM under visual perturbations and achieves encouraging zero-shot generalization, while DINO-WM remains stronger on average. These results suggest that training-time DINO supervision transfers useful visual robustness to the compact latent representation. Although DINO-WM attains a higher average success rate, its planning is approximately $100\times$ slower than \method and LeWM.

\begin{figure}[t]
    \centering
    \includegraphics[width=0.9\textwidth]{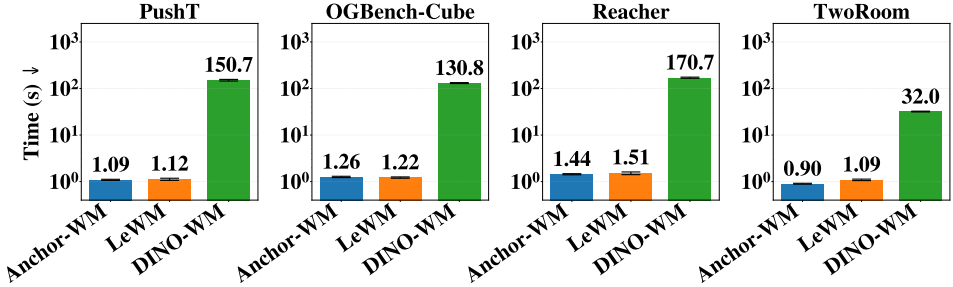}
    \vspace{-3mm}
    \caption{\textbf{Planning Time Comparison.} Anchor-WM achieves planning times comparable to LeWM while being substantially faster than DINO-WM across all four environments.}
    \label{fig:planning_time}
    \vspace{-15pt}
\end{figure}

Additionally, we report planning efficiency in Figure~\ref{fig:planning_time}. Anchor-WM achieves substantially faster planning than DINO-WM, while maintaining planning speed comparable to LeWM. This efficiency stems from performing planning in a compact learned latent space rather than directly over the high-dimensional DINO feature space.

These results show that Anchor-WM combines strong planning performance across standard, long-horizon, and zero-shot settings with efficient inference. By using DINO features as training-time signal and planning in a compact latent space, Anchor-WM retains the benefits of pretrained visual features without incurring the high planning cost of operating directly in the DINO feature space.

%These results show that Anchor-WM can benefit from DINO feature distillation during training to improve planning performance without incurring the high inference-time cost associated with planning directly in the DINO feature space.

\begin{table*}[t]
\centering
\caption{\textbf{Probing the Latent Space.}
Probing results on OGBench-Cube. MSE ($\downarrow$) and Pearson correlation $r$ ($\uparrow$) for linear and MLP probes. Best mean among all methods per metric in bold.}
\label{tab:probe-cube-new}

\resizebox{\textwidth}{!}{
\begin{tabular}{llcccc}
\toprule
Property & Model
& Linear MSE $\downarrow$
& Linear $r$ $\uparrow$
& MLP MSE $\downarrow$
& MLP $r$ $\uparrow$ \\
\midrule

\multirow{3}{*}{Joint Position}
& LeWM
& $0.216 \pm .003$
& $0.702 \pm .002$
& $0.212 \pm .002$
& $0.705 \pm .003$ \\
& DINO-WM
& $0.193 \pm .003$
& $\mathbf{0.736 \pm .002}$
& $0.203 \pm .004$
& $\mathbf{0.741 \pm .004}$ \\
& Anchor-WM (ours)
& $\mathbf{0.192 \pm .003}$
& $0.734 \pm .002$
& $\mathbf{0.192 \pm .003}$
& $0.740 \pm .003$ \\
\midrule

\multirow{3}{*}{Joint Velocity}
& LeWM
& $0.987 \pm .011$
& $0.140 \pm .005$
& $0.997 \pm .011$
& $0.117 \pm .009$ \\
& DINO-WM
& $\mathbf{0.883 \pm .011}$
& $\mathbf{0.315 \pm .004}$
& $0.861 \pm .015$
& $0.357 \pm .003$ \\
& Anchor-WM (ours)
& $0.931 \pm .011$
& $0.254 \pm .005$
& $\mathbf{0.802 \pm .018}$
& $\mathbf{0.427 \pm .006}$ \\
\midrule

\multirow{3}{*}{End-Eff. Position}
& LeWM
& $0.017 \pm .001$
& $0.991 \pm .000$
& $0.003 \pm .000$
& $0.998 \pm .000$ \\
& DINO-WM
& $0.008 \pm .000$
& $0.996 \pm .000$
& $0.003 \pm .000$
& $0.998 \pm .000$ \\
& Anchor-WM (ours)
& $\mathbf{0.006 \pm .000}$
& $\mathbf{0.997 \pm .000}$
& $\mathbf{0.001 \pm .000}$
& $\mathbf{0.999 \pm .000}$ \\
\midrule

\multirow{3}{*}{End-Eff. Yaw}
& LeWM
& $1.000 \pm .014$
& $0.064 \pm .010$
& $1.010 \pm .012$
& $0.093 \pm .018$ \\
& DINO-WM
& $\mathbf{0.893 \pm .012}$
& $\mathbf{0.331 \pm .003}$
& $\mathbf{0.583 \pm .012}$
& $\mathbf{0.647 \pm .005}$ \\
& Anchor-WM (ours)
& $0.902 \pm .012$
& $0.315 \pm .004$
& $0.907 \pm .012$
& $0.316 \pm .010$ \\
\midrule

\multirow{3}{*}{Gripper}
& LeWM
& $\mathbf{0.420 \pm .014}$
& $\mathbf{0.640 \pm .002}$
& $\mathbf{0.388 \pm .014}$
& $\mathbf{0.662 \pm .003}$ \\
& DINO-WM
& $0.528 \pm .014$
& $0.585 \pm .003$
& $0.522 \pm .021$
& $0.594 \pm .007$ \\
& Anchor-WM (ours)
& $0.562 \pm .013$
& $0.563 \pm .003$
& $0.447 \pm .018$
& $0.628 \pm .007$ \\
\midrule

\multirow{3}{*}{Block Position}
& LeWM
& $\mathbf{0.007 \pm .000}$
& $\mathbf{0.996 \pm .000}$
& $\mathbf{0.002 \pm .000}$
& $\mathbf{0.999 \pm .000}$ \\
& DINO-WM
& $0.046 \pm .001$
& $0.976 \pm .001$
& $0.007 \pm .000$
& $0.996 \pm .000$ \\
& Anchor-WM (ours)
& $0.058 \pm .001$
& $0.970 \pm .001$
& $0.004 \pm .000$
& $0.998 \pm .000$ \\
\midrule

\multirow{3}{*}{Block Quaternion}
& LeWM
& $\mathbf{0.919 \pm .116}$
& $0.061 \pm .010$
& $0.935 \pm .106$
& $\mathbf{0.073 \pm .020}$ \\
& DINO-WM
& $0.926 \pm .115$
& $\mathbf{0.065 \pm .007}$
& $0.934 \pm .115$
& $0.050 \pm .007$ \\
& Anchor-WM (ours)
& $0.922 \pm .116$
& $0.044 \pm .009$
& $\mathbf{0.923 \pm .116}$
& $0.035 \pm .010$ \\
\midrule

\multirow{3}{*}{Block Yaw}
& LeWM
& $1.003 \pm .013$
& $0.040 \pm .012$
& $1.021 \pm .011$
& $0.068 \pm .009$ \\
& DINO-WM
& $\mathbf{0.989 \pm .013}$
& $\mathbf{0.134 \pm .007}$
& $1.005 \pm .010$
& $\mathbf{0.114 \pm .003}$ \\
& Anchor-WM (ours)
& $0.999 \pm .012$
& $0.064 \pm .008$
& $\mathbf{1.004 \pm .013}$
& $0.070 \pm .011$ \\
\midrule

\multirow{3}{*}{\textbf{Overall}}
& LeWM
& $0.527 \pm .018$
& $0.492 \pm .002$
& $0.533 \pm .015$
& $0.496 \pm .002$ \\
& DINO-WM
& $\mathbf{0.511 \pm .018}$
& $\mathbf{0.545 \pm .002}$
& $0.513 \pm .019$
& $0.560 \pm .003$ \\
& Anchor-WM (ours)
& $0.526 \pm .019$
& $0.521 \pm .002$
& $\mathbf{0.485 \pm .018}$
& $\mathbf{0.577 \pm .003}$ \\
\bottomrule

\end{tabular}
}
\end{table*}

\subsection{Physical Understanding}
\label{sec:physical_understanding}
\paragraph{Probing Results.}
We probe the latent representations to assess whether they preserve task-relevant physical information. We train linear and MLP probes on frozen latent representations to predict ground-truth physical state variables, and evaluate performance using mean squared error (MSE) and Pearson correlation \(r\). On OGBench-Cube, Anchor-WM preserves substantial information about several task-relevant variables. Its linear probe achieves \(r=0.734\) for joint position and \(r=0.997\) for end-effector position, while the MLP probe reaches \(r=0.998\) for block position. For joint velocity, nonlinear probing improves Anchor-WM from \(r=0.254\) to \(r=0.427\). Across all variables, Anchor-WM achieves an overall linear/MLP correlation of \(0.521/0.577\), compared with \(0.492/0.496\) for LeWM and \(0.545/0.560\) for DINO-WM. Thus, DINO-WM achieves the highest overall linear correlation, while Anchor-WM achieves the highest overall MLP correlation among the evaluated methods. These results suggest that DINO distillation transfers task-relevant physical and geometric information into the compact latent space used for dynamics prediction and planning. Additional probing results are provided in the Appendix.

\begin{figure}[t]
    \centering
    \includegraphics[width=\textwidth]{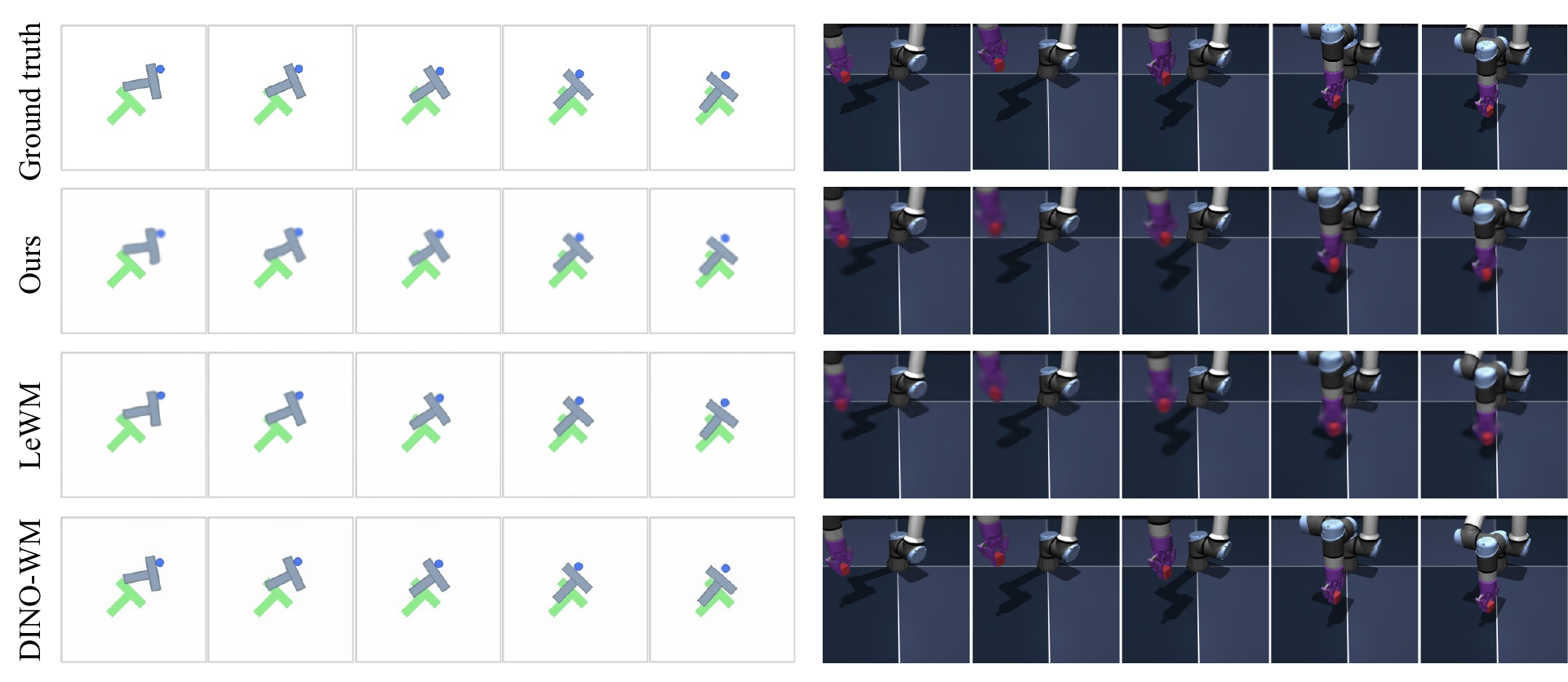}
    \vspace{-15pt}
    \caption{\textbf{Open-Loop Rollouts on PushT and Cube.} DINO-WM produces visually sharper reconstructions, whereas Anchor-WM discards some fine-grained appearance information. Nevertheless, Anchor-WM preserves the task-relevant evolution of object configurations and interactions. Considering the planning results in Table~\ref{tab:success_rate} and Table~\ref{tab:longhorizon}, these examples suggest that visual reconstruction fidelity is not necessarily indicative of a world model’s utility for planning.}
    \label{fig:latent_decoding}
    \vspace{-20pt}
\end{figure}
\paragraph{Latent Space Decoding.} To qualitatively assess the information captured by the learned latent representation, we decode the open-loop predicted latents back into observation space. Figure~\ref{fig:latent_decoding} presents representative examples from PushT and OGBench-Cube. In PushT, the decoded predictions accurately preserve the positions and orientations of task-relevant objects over the rollout and remain closely aligned with the corresponding ground-truth observations. In the visually more complex Cube environment, the predictions likewise retain the overall scene structure and capture the temporal evolution of both the robot and the manipulated object. These results suggest that the compact latent representation preserves task-relevant visual information and that the learned dynamics propagate this information coherently over extended open-loop rollouts.

\subsection{Ablation}
\label{sec:ablations}

\begin{table}[t]
\centering
\caption{\textbf{Ablation Results.} We evaluate the effects of the distillation objective, distillation target, and DINO guidance strategy across all the environments by reporting planning success rate ($\uparrow$).}
\label{tab:ablations}
\small
\setlength{\tabcolsep}{6pt}
\renewcommand{\arraystretch}{1.08}

\begin{tabular}{llcccc}
\toprule
Ablation & Setting & TwoRoom & Reacher & PushT & Cube \\
\midrule

\multirow{2}{*}{Distillation Objective}
& MSE
& 99.6 $\pm$ 0.9 & 82.4 $\pm$ 4.3
& $84.8 \pm 5.4$
& $72.0 \pm 4.2$ \\
& Cosine
& 89.2 $\pm$ 1.8 & 78.6$\pm$ 3.1
& $79.5 \pm 6.8$
& $67.0 \pm 3.8$ \\

\midrule

\multirow{2}{*}{Distillation Target}
& Patch
& 99.6 $\pm$ 0.9 & 82.4 $\pm$ 4.3
& $84.8 \pm 5.4$
& $72.0 \pm 4.2$ \\
& Global
& 83.8 $\pm$ 2.4 & 54.7 $\pm$ 3.7
& $64.3 \pm 3.3$
& $49.6 \pm 5.2$ \\

\midrule

\multirow{2}{*}{DINO Guidance}
& Distillation
& 99.6 $\pm$ 0.9 & 82.4 $\pm$ 4.3
& $84.8 \pm 5.4$
& $72.0 \pm 4.2$ \\
& Alignment
& 50.0 $\pm$ 5.7 & 48.1 $\pm$ 2.3
& $42.7 \pm 5.1$
& $50.4 \pm 3.9$ \\

%\midrule

%\multirow{2}{*}{Encoder Capacity}
%& Dense
%& -- & --
%& $84.8 \pm 5.4$
%& $72.0 \pm 4.2$ \\
%& CLS
%& -- & --
%& $79.5 \pm 6.8$
%& $69.0 \pm 3.8$ \\

\bottomrule
\end{tabular}
\vspace{-15pt}
\end{table}

\paragraph{Distillation Objective Ablation.}
In Eq.~\ref{eq:dino_loss}, we use an MSE objective to match the reconstructed DINO features to the target features. As an alternative, we consider cosine similarity, which measures angular agreement while being insensitive to magnitude, thereby relaxing the constraint imposed by the DINO target and potentially allowing greater flexibility in learning environment-specific dynamics. Table~\ref{tab:ablations} shows that MSE consistently achieves higher planning success rates across all environments. This suggests that preserving both the direction and magnitude of the DINO features provides a more effective training signal than matching directional similarity alone.

\paragraph{Distillation Target Ablation.}
In Eq.~\ref{eq:dino_loss}, the DINO target can be either the spatial patch features or the global CLS representation. We ablate this choice in Table~\ref{tab:ablations}. Using patch-level DINO features yields substantially higher planning success rates across all environments. This suggests that spatially distributed patch features provide richer supervision than a single global representation, helping Anchor-WM preserve task-relevant visual and geometric information during distillation.

\paragraph{DINO Guidance Injection.}
We also study how DINO supervision is incorporated into representation learning. Anchor-WM reconstructs DINO features from the final compact latent through a distillation head, while an alternative following REPA~\citep{yu2025repa} directly aligns layer-8 encoder features with DINO features. Table~\ref{tab:ablations} shows that distillation consistently outperforms intermediate-feature alignment, suggesting that supervising the final latent through a dedicated reconstruction objective is more effective for our model.

We additionally ablate the sensitivity of DINO distillation weights in Fig.~\ref{fig:ablation_lambda} and the influence of latent dimension in Fig.~\ref{fig:ablation_latent_dim} in the Section~\ref{sec:app_result} of the Appendix.

\section{Conclusion and Limitations}

\paragraph{Conclusion.}
We introduced \method, a latent world model that distills pretrained dense DINO representations into a compact latent space for visual planning. This design combines the semantic structure of pretrained visual features with environment-specific dynamics learning, while avoiding representational collapse. \method also allows efficient planning by using the learned compact latent instead of the high-dimensional DINO features. Across the evaluated environments, \method achieves success rates that are competitive with or higher than existing latent world-model baselines and obtains the highest average success rate overall. 

%More broadly, our results highlight a simple alternative to directly operating in large pretrained feature spaces: using pretrained representations as anchors during representation learning, while performing prediction and planning in a compact task-specific latent space.

\paragraph{Limitations.}
\method relies on a pretrained DINO encoder as a training-time teacher and may therefore inherit biases or limitations present in its representation. Exploring alternative pretrained models and more adaptive teacher--student objectives is a promising direction for future work. Furthermore, our evaluation is limited to a relatively small set of offline visual control environments. Extending the evaluation to longer-horizon tasks and real-world robotic settings would provide a more comprehensive assessment of the method.

\section{AI Disclosure}
In this work, we used generative AI to provide feedback on research methodology and experimental design. We did not use generative AI tools to generate synthetic datasets; develop theoretical models or conceptual frameworks; formulate mathematical claims; provide critical ingredients for proving mathematical claims; assist in writing proofs; propose or refine hypotheses; implement methods; assist with translation; clean or reformat datasets; support qualitative or thematic data analysis; or interpret results. Additionally, we used generative AI tools to polish portions of the writing, implement portions of the code, and analyze existing literature. We carefully checked all AI-polished text and reviewed all AI-assisted work. For example, LLM-generated code was independently verified and tested for correctness by two authors. We take full responsibility for the final content of this work, including any text, claims, or artifacts produced with the assistance of generative AI.

\bibliography{iclr2027_conference}

@article{ha2018world,
  title   = {World Models},
  author  = {Ha, David and Schmidhuber, J{\"u}rgen},
  journal = {arXiv preprint arXiv:1803.10122},
  year    = {2018}
}

@article{lecun2022path,
  title   = {A Path Towards Autonomous Machine Intelligence},
  author  = {LeCun, Yann},
  journal = {Open Review},
  year    = {2022}
}

@inproceedings{micheli2023transformers,
  title     = {Transformers are Sample-Efficient World Models},
  author    = {Micheli, Vincent and Alonso, Eloi and Fleuret, Fran{\c{c}}ois},
  booktitle = {The Eleventh International Conference on Learning Representations},
  year      = {2023}
}

@inproceedings{zhou2025dinowm,
  title     = {{DINO-WM}: World Models on Pre-trained Visual Features enable Zero-shot Planning},
  author    = {Zhou, Gaoyue and Pan, Hengkai and LeCun, Yann and Pinto, Lerrel},
  booktitle = {Proceedings of the 42nd International Conference on Machine Learning},
  series    = {Proceedings of Machine Learning Research},
  volume    = {267},
  pages     = {79115--79135},
  year      = {2025},
  publisher = {PMLR}
}

@article{maes2026leworldmodel,
  title   = {{LeWorldModel}: Stable End-to-End Joint-Embedding Predictive Architecture from Pixels},
  author  = {Maes, Lucas and Le Lidec, Quentin and Scieur, Damien and LeCun, Yann and Balestriero, Randall},
  journal = {arXiv preprint arXiv:2603.19312},
  year    = {2026}
}

@article{terver2026ebjepa,
  title   = {A Lightweight Library for Energy-Based Joint-Embedding Predictive Architectures},
  author  = {Terver, Basile and Balestriero, Randall and Dervishi, Megi and Fan, David and Garrido, Quentin and Nagarajan, Tushar and Sinha, Koustuv and Zhang, Wancong and Rabbat, Mike and LeCun, Yann and Bar, Amir},
  journal = {arXiv preprint arXiv:2602.03604},
  year    = {2026}
}

@article{wang2026temporal,
  title   = {Temporal Straightening for Latent Planning},
  author  = {Wang, Ying and Bounou, Oumayma and Zhou, Gaoyue and Balestriero, Randall and Rudner, Tim G. J. and LeCun, Yann and Ren, Mengye},
  journal = {arXiv preprint arXiv:2603.12231},
  year    = {2026}
}

@inproceedings{hafner2019planet,
  title={Learning Latent Dynamics for Planning from Pixels},
  author={Hafner, Danijar and Lillicrap, Timothy and Fischer, Ian
          and Villegas, Ruben and Ha, David and Lee, Honglak
          and Davidson, James},
  booktitle={Proceedings of the 36th International Conference on Machine Learning},
  series={Proceedings of Machine Learning Research},
  volume={97},
  pages={2555--2565},
  year={2019},
  publisher={PMLR}
}

@article{hafner2023dreamerv3,
  title={Mastering Diverse Domains through World Models},
  author={Hafner, Danijar and Pasukonis, Jurgis and Ba, Jimmy
          and Lillicrap, Timothy},
  journal={arXiv preprint arXiv:2301.04104},
  year={2023}
}

@article{hansen2024tdmpc2,
  title={{TD-MPC2}: Scalable, Robust World Models for Continuous Control},
  author={Hansen, Nicklas and Su, Hao and Wang, Xiaolong},
  journal={International Conference on Learning Representations},
  year={2024}
}

@inproceedings{assran2023ijepa,
  title={Self-Supervised Learning from Images with a Joint-Embedding
         Predictive Architecture},
  author={Assran, Mahmoud and Duval, Quentin and Misra, Ishan
          and Bojanowski, Piotr and Vincent, Pascal and Rabbat, Michael
          and LeCun, Yann and Ballas, Nicolas},
  booktitle={Proceedings of the IEEE/CVF Conference on Computer Vision
             and Pattern Recognition},
  pages={15619--15629},
  year={2023}
}

@article{bardes2024vjepa,
  title={Revisiting Feature Prediction for Learning Visual Representations
         from Video},
  author={Bardes, Adrien and Garrido, Quentin and Ponce, Jean
          and Chen, Xinlei and Rabbat, Michael and LeCun, Yann
          and Assran, Mahmoud and Ballas, Nicolas},
  journal={arXiv preprint arXiv:2404.08471},
  year={2024}
}

@inproceedings{caron2021dino,
  title={Emerging Properties in Self-Supervised Vision Transformers},
  author={Caron, Mathilde and Touvron, Hugo and Misra, Ishan
          and J{\'e}gou, Herv{\'e} and Mairal, Julien
          and Bojanowski, Piotr and Joulin, Armand},
  booktitle={Proceedings of the IEEE/CVF International Conference
             on Computer Vision},
  pages={9650--9660},
  year={2021}
}

@article{oquab2023dinov2,
  title={{DINOv2}: Learning Robust Visual Features without Supervision},
  author={Oquab, Maxime and Darcet, Timoth{\'e}e and Moutakanni, Th{\'e}o
          and Vo, Huy V. and Szafraniec, Marc and Khalidov, Vasil
          and Fernandez, Pierre and Haziza, Daniel and Massa, Francisco
          and El-Nouby, Alaa and Assran, Mahmoud and Ballas, Nicolas
          and Galuba, Wojciech and Howes, Russell and Huang, Po-Yao
          and Li, Shang-Wen and Misra, Ishan and Rabbat, Michael
          and Sharma, Vasu and Synnaeve, Gabriel and Xu, Hu
          and J{\'e}gou, Herv{\'e} and Mairal, Julien
          and Labatut, Patrick and Joulin, Armand
          and Bojanowski, Piotr},
  journal={Transactions on Machine Learning Research},
  year={2024}
}

@article{dosovitskiy2020image,
  title={An image is worth 16x16 words: Transformers for image recognition at scale},
  author={Dosovitskiy, Alexey and Beyer, Lucas and Kolesnikov, Alexander and Weissenborn, Dirk and Zhai, Xiaohua and Unterthiner, Thomas and Dehghani, Mostafa and Minderer, Matthias and Heigold, Georg and Gelly, Sylvain and others},
  journal={arXiv preprint arXiv:2010.11929},
  year={2020}
}

@inproceedings{peebles2023scalable,
  title={Scalable diffusion models with transformers},
  author={Peebles, William and Xie, Saining},
  booktitle={2023 IEEE/CVF International Conference on Computer Vision (ICCV)},
  pages={4172--4182},
  year={2023},
  organization={IEEE}
}

@inproceedings{yu2025repa,
  title={Representation Alignment for Generation:
         Training Diffusion Transformers Is Easier Than You Think},
  author={Yu, Sihyun and Kwak, Sangkyung and Jang, Huiwon
          and Jeong, Jongheon and Huang, Jonathan
          and Shin, Jinwoo and Xie, Saining},
  booktitle={International Conference on Learning Representations},
  year={2025}
}

@book{rubinstein2004cross,
  title={The cross-entropy method: a unified approach to combinatorial optimization, Monte-Carlo simulation, and machine learning},
  author={Rubinstein, Reuven Y and Kroese, Dirk P},
  volume={133},
  year={2004},
  publisher={Springer}
}

@inproceedings{nair2023r3m,
  title={{R3M}: A Universal Visual Representation for Robot Manipulation},
  author={Nair, Suraj and Rajeswaran, Aravind and Kumar, Vikash
          and Finn, Chelsea and Gupta, Abhinav},
  booktitle={Proceedings of The 6th Conference on Robot Learning},
  series={Proceedings of Machine Learning Research},
  volume={205},
  pages={892--909},
  year={2023},
  publisher={PMLR}
}

@article{xiao2022mvp,
  title={Masked Visual Pre-training for Motor Control},
  author={Xiao, Tete and Radosavovic, Ilija and Darrell, Trevor
          and Malik, Jitendra},
  journal={arXiv preprint arXiv:2203.06173},
  year={2022}
}

@inproceedings{ma2023vip,
  title={{VIP}: Towards Universal Visual Reward and Representation
         via Value-Implicit Pre-Training},
  author={Ma, Yecheng Jason and Sodhani, Shagun and Jayaraman, Dinesh
          and Bastani, Osbert and Kumar, Vikash and Zhang, Amy},
  booktitle={International Conference on Learning Representations},
  year={2023}
}

@article{assran2025vjepa2,
  title={{V-JEPA 2}: Self-Supervised Video Models Enable
         Understanding, Prediction and Planning},
  author={Assran, Mido and Bardes, Adrien and Fan, David
          and Garrido, Quentin and Howes, Russell and Komeili, Mojtaba
          and Muckley, Matthew and Rizvi, Ammar and Roberts, Claire
          and Sinha, Koustuv and Zholus, Artem and others},
  journal={arXiv preprint arXiv:2506.09985},
  year={2025}
}

@article{hinton2015distilling,
  title={Distilling the Knowledge in a Neural Network},
  author={Hinton, Geoffrey and Vinyals, Oriol and Dean, Jeff},
  journal={arXiv preprint arXiv:1503.02531},
  year={2015}
}

@inproceedings{romero2015fitnets,
  title={{FitNets}: Hints for Thin Deep Nets},
  author={Romero, Adriana and Ballas, Nicolas and Kahou, Samira Ebrahimi
          and Chassang, Antoine and Gatta, Carlo and Bengio, Yoshua},
  booktitle={International Conference on Learning Representations},
  year={2015}
}

@inproceedings{zagoruyko2017attention,
  title={Paying More Attention to Attention:
         Improving the Performance of Convolutional Neural Networks
         via Attention Transfer},
  author={Zagoruyko, Sergey and Komodakis, Nikos},
  booktitle={International Conference on Learning Representations},
  year={2017}
}

@inproceedings{ahn2019variational,
  title={Variational Information Distillation for Knowledge Transfer},
  author={Ahn, Sungsoo and Hu, Shell Xu and Damianou, Andreas
          and Lawrence, Neil D. and Dai, Zhenwen},
  booktitle={Proceedings of the IEEE/CVF Conference on Computer Vision
             and Pattern Recognition},
  year={2019}
}

@inproceedings{tung2019similarity,
  title={Similarity-Preserving Knowledge Distillation},
  author={Tung, Frederick and Mori, Greg},
  booktitle={Proceedings of the IEEE/CVF International Conference
             on Computer Vision},
  year={2019}
}

@inproceedings{radosavovic2023realworld,
  title={Real-World Robot Learning with Masked Visual Pre-training},
  author={Radosavovic, Ilija and Xiao, Tete and James, Stephen
          and Abbeel, Pieter and Malik, Jitendra and Darrell, Trevor},
  booktitle={Proceedings of The 6th Conference on Robot Learning},
  series={Proceedings of Machine Learning Research},
  volume={205},
  pages={416--426},
  year={2023},
  publisher={PMLR}
}

@article{ivashkov2026sensorimotor,
  title={Sensorimotor World Models: Perception for Action via Inverse Dynamics},
  author={Ivashkov, Petr and Balestriero, Randall and Sch{\"o}lkopf, Bernhard},
  journal={arXiv preprint arXiv:2606.20104},
  year={2026}
}

@article{zhang2026deltajepa,
  title={{Delta-JEPA}: Learning Action-Sensitive World Models
         via Latent Difference Decoding},
  author={Zhang, Zhenghao and Wang, Yuanxiang and Guan, Zhenyu
          and Yang, Yujia and Shi, Bingkang and Zong, Tianyu
          and Yi, Hongzhu and Chao, Guoqing and Chen, Xingchen
          and Yang, Tiankun and Bao, Chenxi and Yu, Tao
          and Zhou, Jingjing and Xu, Jungang},
  journal={arXiv preprint arXiv:2606.31232},
  year={2026}
}

@article{bai2026tdjepa,
  title={Temporal-Distance JEPA: Plan-Aware Representation Learning
         for Latent World Model Predictive Control},
  author={Bai, Jiaxin and Xiong, Jiaxuan},
  journal={arXiv preprint arXiv:2607.25337},
  year={2026}
}

@article{m2repa,
  title={Divide and Conquer: Decoupled Representation Alignment
         for Multimodal World Models},
  author={Xiao, Junyuan and Liang, Dingkang and Zhou, Xin
          and Ye, Yixuan and Su, Tongtong and Yi, Guangmo
          and Xia, Bin and Lyu, Qiang and Shi, Shurui
          and Huang, Jun and Si, Jianlou and Yang, Wenming},
  journal={arXiv preprint arXiv:2605.01896},
  year={2026}
}

@article{agra2026making,
  title={Making Foresight Actionable:
         Repurposing Representation Alignment in World Action Models},
  author={Qiu, Lu and Li, Yizhuo and Chen, Yi
          and Ge, Yuying and Ge, Yixiao and Liu, Xihui},
  journal={arXiv preprint arXiv:2606.12217},
  year={2026}
}

@inproceedings{lee2021distill,
  title={Learning to Distill Convolutional Features
         Into Compact Local Descriptors},
  author={Lee, Jongmin and Jeong, Yoonwoo and Kim, Seungwook
          and Min, Juhong and Cho, Minsu},
  booktitle={Proceedings of the IEEE/CVF Winter Conference
             on Applications of Computer Vision},
  pages={898--908},
  year={2021}
}

@article{sobal2022joint,
  title={Joint embedding predictive architectures focus on slow features},
  author={Sobal, Vlad and SV, Jyothir and Jalagam, Siddhartha and Carion, Nicolas and Cho, Kyunghyun and LeCun, Yann},
  journal={arXiv preprint arXiv:2211.10831},
  year={2022}
}

@inproceedings{sobal2025stress,
  title={Stress-testing offline reward-free reinforcement learning: A case for planning with latent dynamics models},
  author={Sobal, Vlad and Zhang, Wancong and Cho, Kyunghyun and Balestriero, Randall and Rudner, Tim GJ and LeCun, Yann},
  booktitle={7th Robot Learning Workshop: Towards Robots with Human-Level Abilities},
  year={2025}
}

@article{bardes2021vicreg,
  title={Vicreg: Variance-invariance-covariance regularization for self-supervised learning},
  author={Bardes, Adrien and Ponce, Jean and LeCun, Yann},
  journal={arXiv preprint arXiv:2105.04906},
  year={2021}
}

@article{balestriero2025lejepa,
  title={Lejepa: Provable and scalable self-supervised learning without the heuristics},
  author={Balestriero, Randall and LeCun, Yann},
  journal={arXiv preprint arXiv:2511.08544},
  year={2025}
}

@inproceedings{hafner2020dream,
  title     = {Dream to Control: Learning Behaviors by Latent Imagination},
  author    = {Hafner, Danijar and Lillicrap, Timothy and Ba, Jimmy and Norouzi, Mohammad},
  booktitle = {International Conference on Learning Representations},
  year      = {2020}
}

@inproceedings{park2025ogbench,
  title={Ogbench: Benchmarking offline goal-conditioned rl},
  author={Park, Seohong and Frans, Kevin and Eysenbach, Benjamin and Levine, Sergey},
  booktitle={International Conference on Learning Representations},
  volume={2025},
  pages={94937--94982},
  year={2025}
}

@article{tassa2018deepmind,
  title={Deepmind control suite},
  author={Tassa, Yuval and Doron, Yotam and Muldal, Alistair and Erez, Tom and Li, Yazhe and Casas, Diego de Las and Budden, David and Abdolmaleki, Abbas and Merel, Josh and Lefrancq, Andrew and others},
  journal={arXiv preprint arXiv:1801.00690},
  year={2018}
}

@article{chi2025diffusion,
  title={Diffusion policy: Visuomotor policy learning via action diffusion},
  author={Chi, Cheng and Xu, Zhenjia and Feng, Siyuan and Cousineau, Eric and Du, Yilun and Burchfiel, Benjamin and Tedrake, Russ and Song, Shuran},
  journal={The International Journal of Robotics Research},
  volume={44},
  number={10-11},
  pages={1684--1704},
  year={2025},
  publisher={Sage Publications Sage UK: London, England}
}
\bibliographystyle{iclr2027_conference}

\appendix
\clearpage
\begin{center}
{\huge Appendix}
\end{center}

In this Appendix, we provide additional experimental results in Section~\ref{sec:app_result}. We further provide details on the dataset used to train \method in Section~\ref{sec:app_data}. Finally, we provide the model architecture, training and planning details of \method in Section~\ref{sec:app_model}.

\section{Additional Results}
\label{sec:app_result}

\begin{table*}[b]
\centering
\small
\setlength{\tabcolsep}{6pt}
\renewcommand{\arraystretch}{1.05}
\caption{Probing results on PushT. MSE ($\downarrow$) and Pearson correlation $r$ ($\uparrow$) for linear and MLP probes. Values are mean $\pm$ standard deviation over five splits. Best of \{Anchor-WM, LeWM\} per metric in bold.}
\label{tab:probe-pusht}
\resizebox{\textwidth}{!}{%
\begin{tabular}{llcccc}
\toprule
Property & Model & Linear MSE $\downarrow$ & Linear $r\uparrow$ & MLP MSE $\downarrow$ & MLP $r\uparrow$ \\
\midrule
\multirow{3}{*}{Agent Location} & LeWM & $0.060\pm.002$ & $0.970\pm.001$ & $\boldsymbol{0.005\pm.001}$ & $\boldsymbol{0.998\pm.001}$ \\
 & DINO-WM & $0.031\pm.002$ & $0.984\pm.001$ & $0.012\pm.003$ & $0.994\pm.001$ \\
 & Anchor-WM (ours) & $\boldsymbol{0.027\pm.002}$ & $\boldsymbol{0.987\pm.001}$ & $0.007\pm.002$ & $0.997\pm.001$ \\
\midrule
\multirow{3}{*}{Block Location} & LeWM & $\boldsymbol{0.028\pm.001}$ & $\boldsymbol{0.986\pm.000}$ & $\boldsymbol{0.001\pm.000}$ & $\boldsymbol{0.999\pm.000}$ \\
 & DINO-WM & $0.026\pm.001$ & $0.987\pm.000$ & $0.008\pm.001$ & $0.996\pm.000$ \\
 & Anchor-WM (ours) & $0.043\pm.002$ & $0.978\pm.001$ & $0.005\pm.001$ & $0.997\pm.000$ \\
\midrule
\multirow{3}{*}{Block Angle} & LeWM & $\boldsymbol{0.173\pm.005}$ & $\boldsymbol{0.909\pm.005}$ & $0.038\pm.004$ & $0.981\pm.002$ \\
 & DINO-WM & $0.216\pm.003$ & $0.885\pm.002$ & $0.036\pm.005$ & $0.982\pm.002$ \\
 & Anchor-WM (ours) & $0.312\pm.009$ & $0.828\pm.004$ & $\boldsymbol{0.032\pm.004}$ & $\boldsymbol{0.984\pm.002}$ \\
\midrule
\multirow{3}{*}{\textbf{Overall}} & LeWM & $\boldsymbol{0.070\pm.001}$ & $\boldsymbol{0.964\pm.001}$ & $\boldsymbol{0.011\pm.001}$ & $\boldsymbol{0.995\pm.001}$ \\
 & DINO-WM & $0.066\pm.001$ & $0.965\pm.000$ & $0.019\pm.001$ & $0.991\pm.001$ \\
 & Anchor-WM (ours) & $0.090\pm.003$ & $0.952\pm.001$ & $0.013\pm.001$ & $0.994\pm.000$ \\
\bottomrule
\end{tabular}%
}
\end{table*}
\paragraph{Additional Probing Results.}
We provide additional probing results on PushT in Table~\ref{tab:probe-pusht}. Anchor-WM preserves substantial information about the agent and object state despite representing each observation with a single latent token. For agent location, Anchor-WM achieves the lowest linear-probe MSE and nearly perfect recovery with the MLP probe ($r=0.997$). For block location, its linear probe is weaker than LeWM and DINO-WM, while the MLP probe reaches $r=0.997$, indicating that the information is readily recoverable through a nonlinear mapping. A similar trend is observed for block angle: although linear probing is relatively weak, the MLP probe reaches $r=0.984$, outperforming both LeWM and DINO-WM. Overall, Anchor-WM improves from $r=0.952$ with the linear probe to $r=0.994$ with the MLP probe, suggesting that the compact latent retains substantial physical information, some of which is encoded nonlinearly.

\begin{figure}[b]
    \centering
    \includegraphics[width=\textwidth]{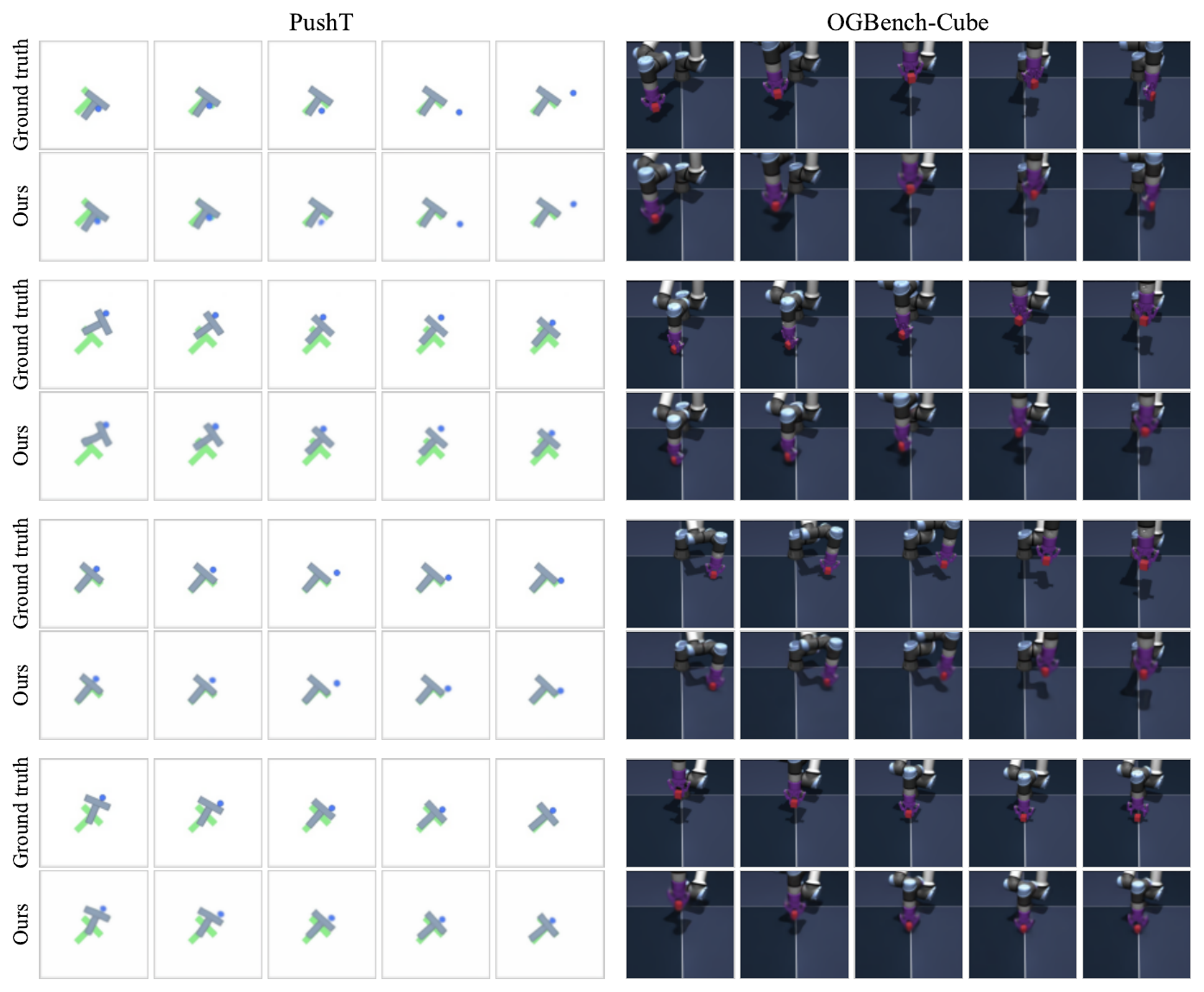}
    \vspace{-15pt}
    \caption{\textbf{Additional open-loop rollout results.} Additional qualitative rollouts on PushT and OGBench-Cube. Each pair of rows compares ground-truth observations with observations decoded from the open-loop latent predictions of \method.}
    \label{fig:appendix_rollouts}
    \vspace{-20pt}
\end{figure}
\paragraph{Additional Open-loop Rollouts.}
Figure~\ref{fig:appendix_rollouts} presents additional open-loop rollout examples for PushT and OGBench-Cube. Across multiple trajectories, the decoded predictions from \method remain qualitatively aligned with the corresponding ground-truth observations over the rollout horizon. In PushT, the predicted latents preserve the motion and orientation of the manipulated object, while in OGBench-Cube they capture the evolution of the robot and cube despite the greater visual complexity of the scene. These examples further illustrate that the learned compact latent representation retains task-relevant visual and dynamical information over open-loop prediction.

\begin{figure}[b]
    \centering
    \includegraphics[width=\textwidth]{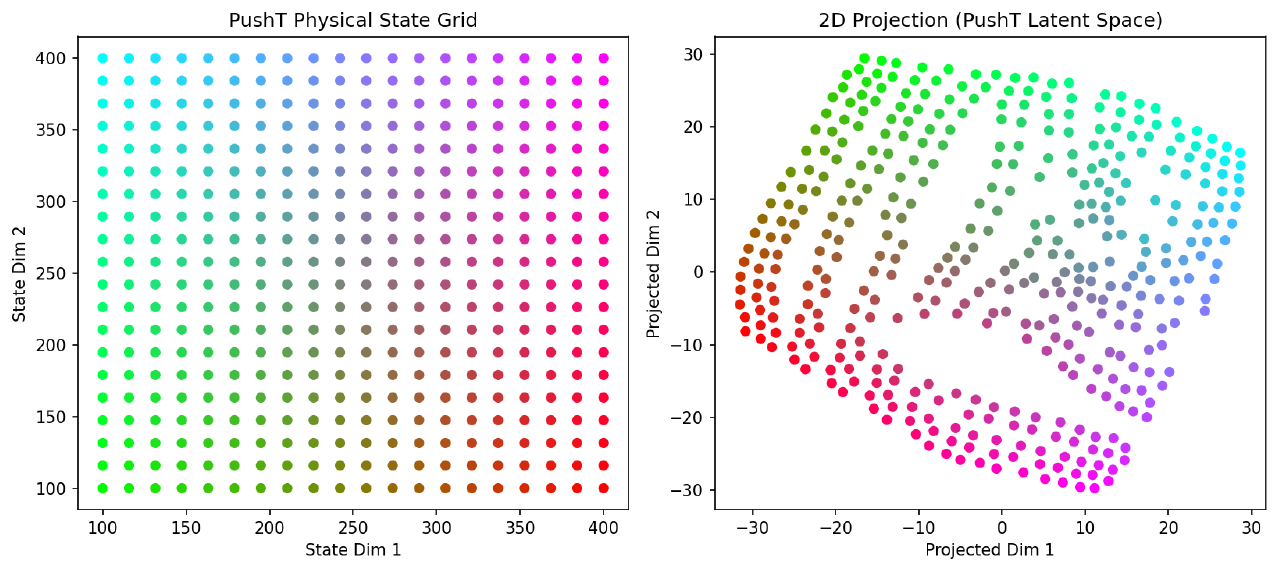}
    \caption{\textbf{Latent-space organization in PushT.} Left: a grid of physical states in PushT, with color indicating the state location. Right: a two-dimensional projection of the corresponding learned latent representations, using the same color assignment. The projected latents exhibit a smooth and coherent organization, with nearby physical states generally remaining close in the learned representation.}
    \label{fig:pca_pusht}
    %\vspace{-1mm}
\end{figure}
\paragraph{Latent Space Visualization.}
We further examine whether the learned latent representation reflects the underlying physical structure of the environment. Figure~\ref{fig:pca_pusht} visualizes a grid of PushT physical states together with a two-dimensional projection of their corresponding latent representations. The projected latents form a structured and smoothly varying manifold: neighboring physical configurations tend to appear near one another, and changes in physical state are accompanied by gradual transitions in the latent space. This qualitative organization suggests that \method preserves meaningful geometric relationships between physical states rather than mapping observations to an unstructured representation.

\begin{figure}[t]
    \centering
    \includegraphics[width=0.6\textwidth]{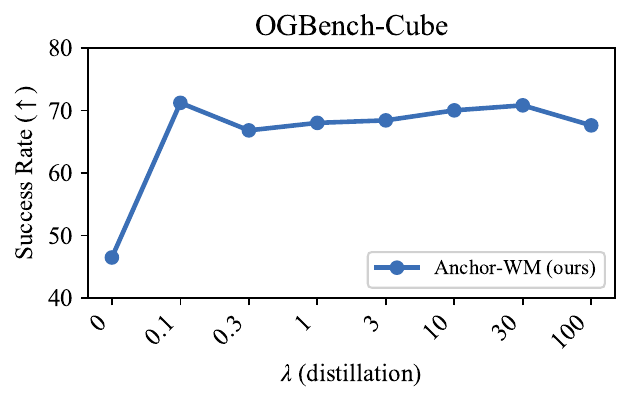}
    \vspace{-3mm}
    \caption{\textbf{Effect of the distillation weight $\lambda$.} Planning success rate of \method on OGBench-Cube as the weight $\lambda$ of the DINO distillation objective is varied. Setting $\lambda=0$ removes the distillation objective. Performance improves substantially when distillation is introduced and remains relatively stable across a broad range of nonzero values, indicating that \method is not highly sensitive to the precise choice of $\lambda$.}
    \label{fig:ablation_lambda}
    \vspace{-1mm}
\end{figure}
\paragraph{Sensitivity to the Distillation Weight.}
We study the sensitivity of \method to the weight $\lambda$ applied to the DINO distillation objective on OGBench-Cube. As shown in Fig.~\ref{fig:ablation_lambda}, removing the distillation objective ($\lambda=0$) leads to a substantial drop in planning success rate. In contrast, performance improves markedly for nonzero values of $\lambda$ and remains relatively stable over a broad range of settings. This indicates that the benefit of DINO supervision does not depend on carefully tuning the distillation weight, and that \method is reasonably robust to the choice of $\lambda$.

\begin{figure}[t]
    \centering
    \includegraphics[width=0.6\textwidth]{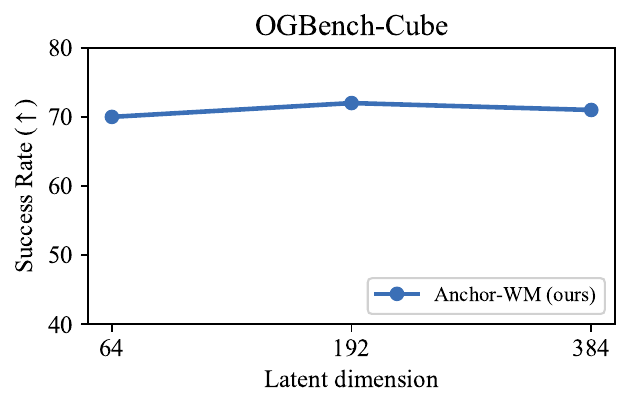}
    \vspace{-3mm}
    \caption{\textbf{Effect of latent dimensionality.} Planning success rate of \method on OGBench-Cube with latent dimensions of 64, 192, and 384. Performance remains similar across all three settings, indicating that \method is relatively insensitive to the latent dimensionality within this range.}
    \label{fig:ablation_latent_dim}
    \vspace{-1mm}
\end{figure}
\paragraph{Sensitivity to Latent Dimensionality.}
We evaluate the effect of the compact latent dimensionality on planning performance in OGBench-Cube. As shown in Fig.~\ref{fig:ablation_latent_dim}, performance varies only slightly when the latent dimension is changed from 64 to 384, with the 192-dimensional representation achieving the highest success rate. Increasing the latent size to 384 does not provide additional improvement, while even a 64-dimensional latent retains similar performance. These results suggest that the planning performance of \method is not highly sensitive to the latent dimensionality within the evaluated range and that strong performance can be achieved with a compact representation.

\begin{figure}[t]
    \centering
    \includegraphics[width=\textwidth]{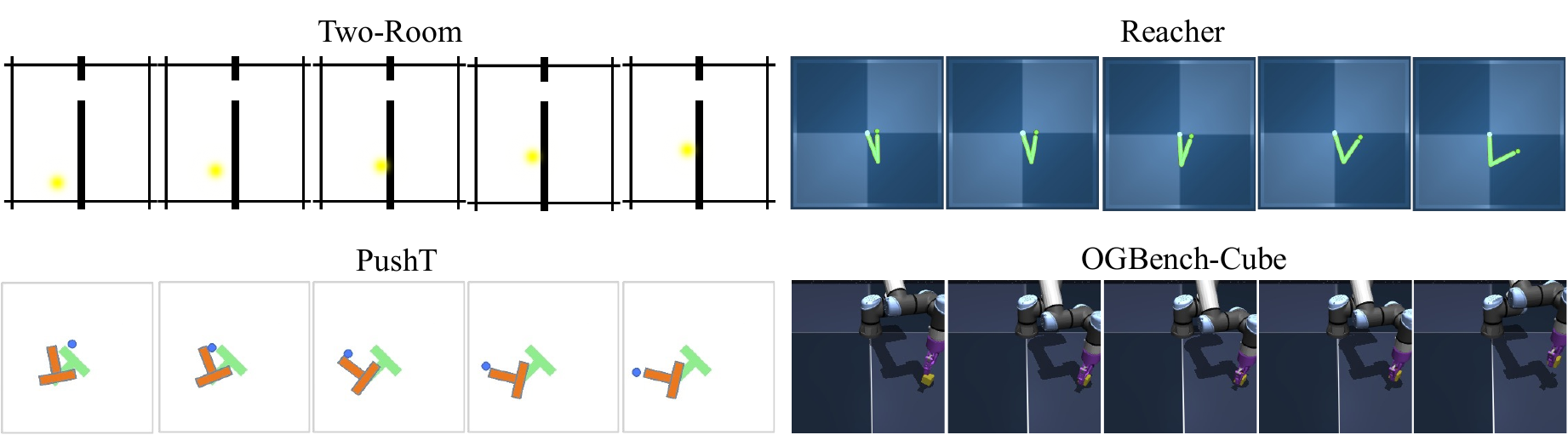}
    \caption{\textbf{Out-of-Distribution Environment Illustration.} Top Left: Two-Room, we change the agent color from red to yellow. Top-Right: Reacher, we change the arm color from blue to green.  Bottom Left: Push-T, we change the object color from grey to orange. Bottom Right: OGBench-Cube, we change the cube color from red to brown.}
    \label{fig:environments_ood}
    %\vspace{-1mm}
\end{figure}

\section{Datasets}
\label{sec:app_data}
We evaluate Anchor-WM on a set of continuous-control environments spanning navigation and robotic manipulation. For every environment, we reserve 10\% of the collected trajectories as a held-out test set and use the remaining 90\% for training. Our evaluation includes both in-distribution (ID) settings, where test trajectories follow the same environment configuration and data distribution as the training data, and out-of-distribution (OOD) settings, where the model is evaluated under modified visual or task conditions not observed during training.

\paragraph{TwoRoom.}
TwoRoom is a continuous 2D navigation environment consisting of two regions connected by a narrow doorway. The agent must move from an initial position to a goal located in the opposite room, making successful trajectories depend on navigating through the doorway. We use a dataset of 10,000 trajectories generated by a noisy waypoint-based policy that first guides the agent toward the doorway and subsequently toward the sampled goal.

\paragraph{PushT.}
PushT is a planar manipulation task in which a point-like agent interacts with a T-shaped object through pushing actions. The objective is to move and orient the object so that it matches a desired target configuration. We use the standard PushT dataset containing 20,000 expert demonstrations, following the data setup commonly used in prior visual world-model work.

\paragraph{OGBench-Cube.}
OGBench-Cube is a 3D robotic manipulation task in which a robot arm must grasp a cube and transport it to a specified target position. We use the single-object version of the environment and collect 10,000 trajectories of 200 environment steps using the benchmark's provided data-collection policy.

\paragraph{Reacher.}
Reacher is a continuous-control task from the DeepMind Control Suite in which a two-link arm must position its end effector at a target location in the plane. We use 10,000 trajectories of 200 steps collected with a Soft Actor-Critic policy. Following prior world-model evaluations, success requires the arm to reach the target configuration with high precision.

\paragraph{ID and OOD evaluation.}
The held-out 10\% split is used for ID evaluation, providing unseen trajectories from the same data-generating process as the training set. OOD evaluation instead measures performance under visual perturbation as illustrated in Fig.~\ref{fig:environments_ood}, so that the model is tested on observations or task configurations that differ from those encountered during training.

\section{Architecture, Training, and Planning}
\label{sec:app_model}

We use a Vision Transformer (ViT-Tiny, patch size 14; \citep{dosovitskiy2020image}), trained from scratch, to parameterize the Anchor-WM encoder. Each $224\times224$ observation is mapped to a single 192-dimensional latent representation given by the \texttt{[CLS]} token. Actions are $z$-scored independently in each dimension, grouped over the frame-skip interval of 5 environment steps, and embedded into the same 192-dimensional space.

The dynamics model is a 6-layer causal Transformer with 16 attention heads and an MLP dimension of 2048. It is conditioned on the action embeddings using adaptive layer normalization (AdaLN-zero; \citep{peebles2023scalable}). Given a history of 3 latent frames, the model autoregressively predicts the next latent representation and is trained using a mean-squared error loss on the predicted next embedding.

To anchor the learned latent space, we additionally train a 2-layer MLP head to reconstruct the final-layer patch tokens of a frozen DINOv2 ViT-S/14 teacher \citep{oquab2023dinov2}. The teacher features are computed from images at $196\times196$ resolution, and their reconstruction from the Anchor-WM latent is optimized with a mean-squared error loss weighted by $\lambda$. This auxiliary objective replaces the anti-collapse regularizer used in LeWM \citep{maes2026leworldmodel}. The DINOv2 teacher and reconstruction head are used only during training and discarded at inference, so planning relies only on the compact encoder and dynamics predictor.

We learn a decoder only for visualization purpose.
Anchor-WM is trained without any pixel reconstruction, so to visualize its latents we train a separate
decoder post hoc, after world-model training. The world-model encoder is frozen and kept in evaluation
mode; only the decoder is optimized, and it is never used during world-model training or planning. The
decoder maps the 192-dimensional latent to a $224\times224$ RGB image: a linear layer projects the latent
to a $256\times7\times7$ feature map, followed by five upsampling stages, each consisting of bilinear
$2\times$ upsampling and two $3\times3$ convolutions with group normalization and GELU activations
(channels $256\!\to\!128\!\to\!64\!\to\!32\!\to\!16\!\to\!16$), and a final $3\times3$ convolution with a
sigmoid output ($\sim$3.0M parameters in total). It is trained with a pixel-wise mean-squared error on RGB
values in $[0,1]$ using AdamW (learning rate $3\times10^{-4}$, weight decay $10^{-4}$, batch size 64) for
20{,}000 steps on frames sampled every 5 environment steps from the training data, and we keep the
checkpoint with the lowest reconstruction error on 1{,}024 frames from episodes held out from decoder
training. For open-loop rollouts, the encoder embeds the context frames, the predictor rolls the latent
forward under the ground-truth actions, and each predicted latent is decoded independently. For
comparison, LeWM decoders are trained with the identical architecture and protocol.

All world models are evaluated using the same planner and evaluation protocol. We use the cross-entropy method (CEM) with 300 sampled action sequences, 30 optimization iterations, and the top 30 samples as elites. Each latent step corresponds to a block of 5 environment actions (frame skip 5), and plans span 5 latentsteps, i.e., 25 environment steps. We execute the full planned sequence and then replan from the currentobservation (receding horizon of 5 latent steps), resulting in two planning calls per 50-step episode. Planning is performed in a receding-horizon manner from the current observation. The planning objective minimizes the squared distance between the final predicted latent and the latent representation of the goal observation.

For standard planning evaluation, the goal is selected 25 environment steps after the start state along an expert trajectory, and each episode has a budget of 50 environment steps. For long-horizon evaluation, the goal is selected 50 environment steps ahead and the budget is increased to 100 environment steps. Start--goal pairs are sampled from the 10\% of episodes held out from training. An episode is considered successful if the agent reaches the goal within the evaluation budget. We report the mean success rate over 5 random seeds, with 50 evaluation episodes per seed. We additionally report planning time as a measure of inference efficiency, with all methods evaluated on a single NVIDIA A100 GPU.

The model, training, and planning hyperparameters of Anchor-WM are summarized in Table~\ref{tab:hparams}.
 
\begin{table}[t]
\centering
\caption{Anchor-WM hyperparameters (shared across environments).}
\label{tab:hparams}
\begin{tabular}{ll}
\toprule
\textbf{Hyperparameter} & \textbf{Value} \\
\midrule
Encoder & ViT-Tiny (patch size 14), from scratch \\
Input resolution & $224 \times 224$ \\
Latent dimension & 192 (CLS token) \\
Predictor & Causal Transformer, 6 layers \\
Attention heads / head dim & 16 / 64 \\
Predictor MLP dimension & 2048 \\
Predictor dropout & 0.1 \\
Action conditioning & AdaLN-zero \\
History length & 3 frames \\
Frame skip & 5 \\
Distillation teacher & DINOv2 ViT-S/14 (frozen) \\
Teacher input resolution & $196 \times 196$ \\
Distillation target & Last-layer patch tokens \\
Distillation head & 2-layer MLP (GELU) \\
Distillation loss & MSE \\
Optimizer & AdamW \\
Learning rate & $5 \times 10^{-5}$ \\
Weight decay & $10^{-3}$ \\
Batch size & 128 \\
Precision & bfloat16 \\
Gradient clipping & 1.0 \\
Planner & CEM \\
CEM samples / iterations / elites & 300 / 30 / 30 \\
Planning horizon & 5 \\
%Receding horizon (MPC) & 5 \\
Evaluation episodes & 50 \\
Goal offset / step budget & 25 / 50 \\
\bottomrule
\end{tabular}
\end{table}

\clearpage
\newpage

\end{document}